\documentclass[10pt,twocolumn,letterpaper]{article}

\usepackage{cvpr}
\usepackage{times}
\usepackage{epsfig}
\usepackage{graphicx}
\usepackage{amsmath}
\usepackage{amssymb}
\usepackage{mathtools}
\usepackage{algorithm}
\usepackage{algorithmic}

\usepackage[pagebackref=true,breaklinks=true,letterpaper=true,colorlinks,bookmarks=false]{hyperref}
\usepackage{multirow}

\cvprfinalcopy
\def\cvprPaperID{19}

\ifcvprfinal\fi

\begin{document}

\title{On-Device Neural Net Inference with Mobile GPUs}

\author{
Juhyun Lee,
Nikolay Chirkov,
Ekaterina Ignasheva,
Yury Pisarchyk,
Mogan Shieh, \\
Fabio Riccardi,
Raman Sarokin,
Andrei Kulik, and
Matthias Grundmann\\
Google Research\\
1600 Amphitheatre Pkwy, Mountain View, CA 94043, USA\\
{\tt\small \{impjdi,chirkov,eignasheva,ypisarchyk,moganshieh,fricc,sorokin,akulik,grundman\}@google.com}
}

\maketitle

\begin{abstract}
   On-device inference of machine learning models for mobile phones is desirable
   due to its lower latency and increased privacy.  Running such a
   compute-intensive task solely on the mobile CPU, however, can be difficult
   due to limited computing power, thermal constraints, and energy consumption.
   App developers and researchers have begun exploiting hardware accelerators to
   overcome these challenges.  Recently, device manufacturers are adding neural
   processing units into high-end phones for on-device inference, but these
   account for only a small fraction of hand-held devices. In this paper, we
   present how we leverage the mobile GPU, a ubiquitous hardware accelerator on
   virtually every phone, to run inference of deep neural networks in real-time
   for both Android and iOS devices.  By describing our architecture, we also
   discuss how to design networks that are mobile GPU-friendly.  Our
   state-of-the-art mobile GPU inference engine is integrated into the
   open-source project TensorFlow Lite and publicly available at
   \small{\url{https://tensorflow.org/lite}}.
\end{abstract}

\section{Introduction}

On-device machine learning (ML) offers a variety of benefits. The most apparent
is the improved inference latency: By skipping the data upload to the server and
wait-time for the inference result, the app can respond more quickly to the
user's request.  Removing the server dependency has additional benefits, such
as:
\begin{itemize}
\setlength{\parskip}{0pt}
\setlength{\itemsep}{0pt plus 1pt}
\item Removing the need to maintain inference servers,
\item Running with limited or no connectivity, and
\item Reducing privacy concerns as the user data remains on the device.
\end{itemize}

However, on-device ML is not trivial. Despite both recent advances in mobile
hardware technology and efforts to efficiently run deep networks on mobile
devices, mobile CPUs continue to be less powerful than those found in servers.
Running deep net inference on a mobile device means adding a significant
compute-intensive task to the CPU which competes with existing logic.  Fully
utilizing the mobile CPU comes with additional unwanted costs, \eg increased
energy consumption leads to shorter battery life and an increase in the phone's
thermal profile causes throttling resulting in slower computation.

Hardware accelerators such as the digital signal processors offer solutions
to overcome these challenges.  The demand for on-device ML has led to recent
trends of phone manufacturers integrating dedicated neural processing
units (NPUs) for high-end next-generation phones, which account for only
a small fraction of the current distribution of mobile devices.

Our primary goal is a fast inference engine with wide coverage for TensorFlow
Lite (TFLite) \cite{tflite}.  By leveraging the mobile GPU, a ubiquitous
hardware accelerator on virtually every phone, we can achieve real-time
performance for various deep network models. Table \ref{cpugpu} demonstrates
that GPU has significantly more compute power than CPU.

\begin{table}[!h]
\begin{center}
\begin{tabular}{|l|c|c|}
\hline
Device & CPU (FP32) & GPU (FP16)\\
\hline\hline
Samsung Galaxy S5 & 79 & 300 \\
Samsung Galaxy S7 & 124 & 730 \\
Samsung Galaxy S9 & 270 & 730 \\
\hline
\end{tabular}
\end{center}
\caption{
  Example of available compute power on mobile in gigaflops (billion floating
  point instructions per second).  FP16 and FP32 refer to 16- and 32-bit
  floating point arithmetic, respectively.}
\label{cpugpu}
\end{table}

This paper presents the techniques we adopt for TFLite GPU and how we achieve an
average acceleration of 2--9$\times$ for various deep networks on GPU
compared to CPU inference.  We first describe the general mobile GPU
architecture and GPU programming, followed by how we materialize this with
Compute Shaders for Android devices, with OpenGL ES 3.1+ \cite{leech2016opengl}
and Metal Shaders for iOS devices with iOS 9+ \cite{apple2014metal}.

\section{Related Work}

Various research efforts from both academia and industry endeavor to bring deep
neural networks inference previously limited to server, forward to mobile
devices.  Those efforts can be roughly categorized into three strategies: 
\begin{itemize}
\setlength{\parskip}{0pt}
\setlength{\itemsep}{0pt plus 1pt}
\item Network architecture-driven, 
\item Hardware-driven, and 
\item ML framework-driven.
\end{itemize}

Neural network researchers have focused on optimizing their network
architectures explicitly for processing on-device in various domains such as
image classification~\cite{howard2017mobilenets,sandler2018mobilenetv2},
object localization~\cite{huang2017speed}, and image enhancements
~\cite{ignatov2017dslr,ignatov2018pirm}.  Many of these techniques involve
reducing the model size by re-designing the network architecture and adding
pre-/post-training quantization of weights.  With these, one can achieve
faster computation and smaller memory footprint, leading to reduced
inference latency at the cost of slightly degraded model accuracy.
MorphNet~\cite{gordon2018morphnet} takes a unique path of reducing
the number of floating point operations per second which is optimized during
training of the model.
Our work is complementary to these efforts and instead focuses on optimizing
the inference engine that runs the neural network rather than the model or
training.

Major hardware manufacturers have made architectural changes responding to
demands for faster mobile inference, and are publishing software development
kits (SDKs) to expose those:
Arm Compute Library~\cite{arm-compute-library},
Huawei HiAI SDK~\cite{hiai},
MediaTek NeuroPilot SDK~\cite{neuropilot}, and
Qualcomm SNPE SDK~\cite{snpe}.
These libraries are vendor-specific and either cannot be re-used on a
different architecture or do not guarantee the expected performance boost on
other platforms.  Our work does not add new hardware or SDKs.  Instead, we
use well-established hardware, the mobile GPU, and well-supported graphics
and compute standards as OpenGL~\cite{leech2016opengl} and Metal
\cite{apple2014metal}, to achieve high-performance neural network inference.

Apple presented the Metal
Performance Shaders with support of convolutional neural networks \cite{mpscnn}
accelerated by GPU.  This is a solution built on top of the Metal API and
allows custom operations.  Our approach is analogous to Apple's on iOS devices.
Apple also released CoreML \cite{coreml}, an end-to-end solution for inference
on mobile devices using CPU, GPU, and NPU, if available.

Android introduced the Android Neural
Networks API \cite{nnapi} that serves as a layer between hardware and
higher-level ML frameworks that vendors must implement for Android 8.1 or later.
Our work has wider coverage and does not depend on a specific Android version,
or require vendors to implement individual APIs for deep network processing.

Some of the latest mobile-friendly ML frameworks are:
\begin{itemize}
\setlength{\parskip}{0pt}
\setlength{\itemsep}{0pt plus 1pt}
\item Caffe2~\cite{caffe2} which focuses on CPU inference and
      uses Arm Compute Library for Arm Mali GPUs.
\item MACE~\cite{mace} which employs OpenCL which is not a part of standard
      Android OS.
\end{itemize}
TFLite GPU leverages the mobile GPU with OpenGL ES for Android devices and
Metal for iOS devices.  The specific version requirements are OpenGL ES 3.1+
and iOS 9+ which are available for more than 52\% of all Android devices
\cite{wu19}.
One of our biggest strength is that our framework employs open standards, \ie is
not limited by specific hardware vendor, and thus covers a wide range of
devices.

\section{General Architecture}

This section explains the general architecture of TFLite GPU, consisting of an
initialization phase followed by a model inference phase.  The techniques
in this section are independent of the architecture of the underlying
GPU.

\subsection{Initialization}

\begin{figure}[b]
\begin{center}
   \includegraphics[height=4.2cm]{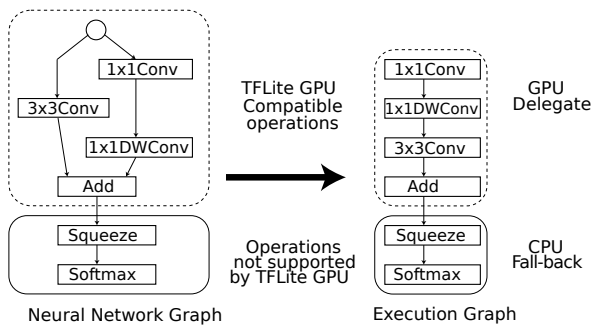}
\end{center}
\caption{
  TFLite's delegate mechanism: Operations supported by the GPU delegate will run
  on the GPU, and the rest on the CPU.
}
\label{fig:delegate}
\end{figure}

TFLite provides APIs for the delegation of the  execution of neural network
sub-graphs to another library.  We exploit this feature to integrate the GPU
backend into TFLite.  Given a neural net model, TFLite first checks whether it
can execute all the operators in the model with our GPU delegate.  Our GPU
backend identifies supported operators, and TFLite then partitions the  graph
into several sub-graphs, substituting the sub-graphs with virtual ``delegate
nodes''.  From that point, the GPU backend is responsible for executing this
sub-graph, as depicted in Figure~\ref{fig:delegate}.  Unsupported operators are
by default computed by the CPU.  Ideally, the whole graph would be compatible
with our mobile GPU backend for maximum performance.  

As our mobile GPU inference engine is primarily designed for high-performance
execution, we first inspect the model and resolve obvious inefficiencies. For
example:
\begin{itemize}
\setlength{\parskip}{0pt}
\setlength{\itemsep}{0pt plus 1pt}
\item Merging \textsc{pad} as an option of another op
      where it was previously described separately.
\item Removing superfluous identity operations, \eg \textsc{resize} with scale
      one or single input \textsc{add/concat}.
\end{itemize}
While these inefficiencies might be caught by the architect, artifacts such as
these crop up inevitably, and we should still optimize these whenever possible.

Note that, in contrast to CPU backends which work without initialization,
GPU backends require initialization involving shader compilation and
optimization by the driver before inference. The cost of this process depends on
network size and may take from few milliseconds to seconds, but is incurred once
and not again for subsequent runs until the cache memory is invalidated for any
of reasons: application is updated or re-installed, device is rebooted, cache
memory is over, or for other OS-specific reasons.

\subsection{Running Inference}

The inference phase is fairly straightforward.  The input tensors are reshaped
to the PHWC4 format detailed later in Section~\ref{phwc4}, if their tensor shape
has channel size not equal to 4.  For each operator, shader programs are linked
by binding resources such the operator's input/output tensors, weights, \etc.
and dispatched, \ie inserted into the command queue.  The GPU driver then takes
care of scheduling and executing all shader programs in the queue, and makes
the result available to the CPU by the CPU/GPU synchronization. There might be a
final conversion from PHWC4 to HWC, if the output tensor has a channel size not
equal to 4.

For maximum performance, one should avoid CPU/GPU synchronization at all cost,
and preferably, never leave GPU context if real-time processing is needed.  The
most ideal scenario would be the following: A camera provides with RGBA texture
that goes directly to TFLite GPU and the output of the network is then directly
rendered to the screen.

\vspace{-\topsep}
\paragraph{Shader Program Optimization}
In the GPU inference engine, operators exist in the form of shader programs.
The shader programs eventually get compiled and inserted into the command 
queue and the GPU executes programs from this queue without synchronization 
with the CPU. 

To reduce the number of shader programs in the command queue, we consolidate
them into meaningful aggregates while maximizing parallelism and well-defined
data dependencies.  The following techniques are employed when generating the
source code for the shader programs:
\begin{itemize}
\setlength{\parskip}{0pt}
\setlength{\itemsep}{0pt plus 1pt}
\item Fusing element-wise operators with computationally expensive operators,
      \eg activations with convolution, to reduce the number of shader programs.
\item In-lining parameters and small objects directly into the shader program to
      reduce memory I/O overhead. 
\item Baking uniforms into the source code, instead of passing them in the
      run-time, allowing drivers to produce more optimal code.
\item Creating specialized version of shaders, like ``convolution with
      $1{\times}1$ kernel size'', to manually optimize shaders for particular
      cases.
\item Implementing specialization of shader programs optimized for a certain
      architecture to improve the op's performance on the said environment.
\end{itemize}

After the source code for each program is generated, each shader gets compiled.
This compilation step can take a while, from several milliseconds to seconds.
Typically, app developers can hide this latency while loading the
model or starting the app for the first time. Once all shader programs are
compiled, the GPU backend is ready for inference.

\section{Data Layout}
\label{phwc4}

\begin{figure}[t]
\begin{center}
   \includegraphics[height=4cm]{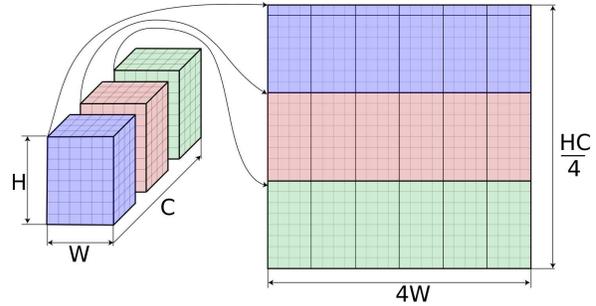}
\end{center}
\caption{
  Example of PHWC4 memory layout (best viewed in color).  A tensor of shape
  $(H{=}8, W{=}6, C{=}12)$ is split into 4-element slices of size
  $(H, W, 4)$ which are stored sequentially as a continuous 2D array of size
  $(HC/4{=}24, 4W{=}24)$.}
\label{fig:phwc4}
\end{figure}

Most modern GPUs use a homogeneous coordinate~\cite{moebius1827der} system which
represents points in space with coordinates $(x,y,z,w)$.  A homogeneous
coordinate $(x,y,z,w)$, where $w{\neq}0$, represents a point $(x/w,y/w,z/w,1)$
in a 3D space.  This allows affine transformations and projective
transformations to be represented in the form of 4D matrix multiplications.
GPUs are essentially processors optimized for 4-element vector compute and
load/store operations.

While TFLite does not restrict tensors to a certain shape, many operators assume
4D input/output tensors shaped as $[B,H,W,C]$ where $B$, $H$, $W$, $C$
respectively represent batch size, height, width, and channel size.  For
convenience, the rest of the paper will mostly describe tensors assuming a batch
size of $1$, or $[H,W,C]$ for short.  This simplified example can be generalized
if we consider batches to be a concatenation of multiple $[H,W,C]$ tensors.

In TFLite GPU, a $[H,W,C]$ tensor is split into 4-channel slices which are
stored sequentially in memory.  If the number of channels is not divisible by
$4$, it is padded with zeroes.  This memory layout, called PHWC4 (Figure
\ref{fig:phwc4}), optimally reduces cache misses in the graphics architecture.
This is tightly coupled with how compute threads are executed on the GPU, which
defines the order of computation, and more importantly, the order of memory load
instructions.

\begin{figure}[t]
\begin{center}
   \includegraphics[height=3.4cm]{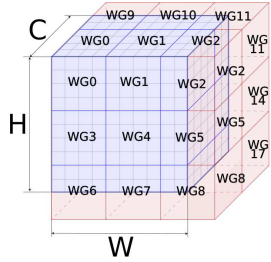}
\end{center}
\caption{
  Compute shader execution grid $(X{=}12,Y{=}12,Z{=}8)$ built upon the tensor
  shape $(H{=}10,W{=}10,C{=}6)$ shown in blue (best viewed in color). Work group
  size $(x{=}4,y{=}4,z{=}4)$ highlighted as cubes with bold lines. Each cell
  represents a FP32 value.
}
\label{fig:compute}
\end{figure}

\subsection{Work Groups: GPU Threading Units}

A GPU compute task consist of a shader program and a grid.  Every thread
executes the same shader program, but on different region of a 3D mesh problem
space.  The global grid is made up of repeated work groups of constant shape
$(x,y,z)$ and has a total dimension $(X,Y,Z)$ which is a multiple of these work
groups.

Every operation in the graph has at least one output 3D tensor.  If there is
more than one output tensor, we use one of them as a basis for the compute grid
size calculation. The grid may be larger than the actual output tensor, because
we expand it to sizes in multiples of 4 due to GPUs working efficiently for
those sizes. This causes the creation of threads which do nothing and return at
the beginning of the main function, but this is faster than working with
misaligned grid sizes which prevents efficient optimization of byte code. The
described situation is visualized in Figure~\ref{fig:compute}, where blue color
highlights useful threads which will actually calculate output values, and red
color highlights stub threads. Further tuning of the compute grid/work group
sizes is described in subsection~\ref{wg_picker}.

Optimizations are focused on neighboring threads \textit{within} a work group -
those spawned in sequential order as described. The PHWC4 layout provides the
advantage of allowing neighboring threads to hit the same cache line when
requesting data for input tensors.

Threads inside a work group are executed in a particular order. Our experiments
show that for each work group channel, each row is sequentially picked in order
from the first to last, starting across $W$, then $H$ and finally $C$. Ordering
of work group execution is likewise sequential and follows the same schema, as
shown on Figure~\ref{fig:compute}.

\begin{figure}[t]
\begin{center}
   \includegraphics[height=4cm]{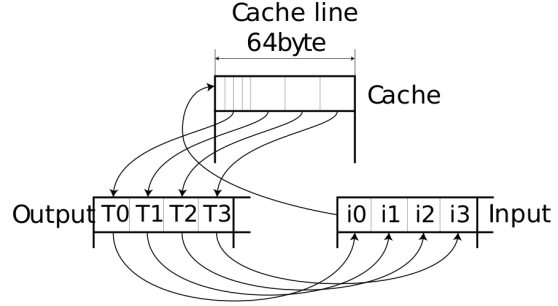}
\end{center}
\caption{
  Cache hit by 4 neighboring threads.  When threads $T_0$--$T_3$ each issue
  a 16-byte load of memory blocks $i_0$--$i_3$ that are contiguous in memory,
  the first load can fill the 64-byte cache line, benefiting the other threads
  with no additional cost in memory I/O.
}
\label{fig:cache}
\end{figure}

\vspace{-\topsep}
\paragraph{For a 2D Convolution,} we compute the result at every output element, by iterating over the weights of a convolution kernel and its corresponding input elements covered by a window of size $(\mathit{kernel\_height}, \mathit{kernel\_width})$. For simplicity, we consider the case of $1{\times}1$ convolution window case. In this case, only one input cell is needed to calculate one output element. As we work with 3D tensors, every cell is implied to be a vector of channels. For this operation, every thread at the very first iteration of its loop requests first 4 channels of the appropriate cell. A compulsory cache miss occurs on the initial thread request (for 16 bytes, or 4 float values), which triggers the actual data load. When this occurs, the hardware memory manager loads the whole cache line and not just the requested 16 bytes.  Since the cache line size on most mobile GPUs is 64 bytes, this results in the loading of the next 48 bytes as well. Since all threads execute the same shader code, the neighboring threads will also execute the same code as the first one (the initially requested 16 bytes). Organizing threads in the way is an efficient strategy for memory loading as the next (neighboring) input values will already be available when requested and loaded as part of the same cache line for initial neighbor compute threads (Figure~\ref{fig:cache}).  

\subsection{Work Group Size Selection}
\label{wg_picker}

The work group size for executing shader programs defines the group of threads
which share data inside the work group.  Depending on the GPU, picking the right
work group size can result in increased performance, whereby picking the wrong
can result in unexpected slowdowns.  Arm Mali GPUs, for instance, show robust
performance independent of configured work group sizes and tuning them only
results in a nominal performance gain typically less than 5\%.  Qualcomm Adreno
GPUs, on the other hand, are extremely sensitive to well-configured work group
sizes and tuning these can give up to a 30\% performance boost.

Tuning the work group size is unfortunately difficult as GPU internals are not
available to the user either directly (via the API), or indirectly (via some
assembly representation of internal state).  Threads are executed in groups
called ``waves'' and knowing the wave size is crucial to optimizing the work
group size as they fine-tune the memory usage of neighboring threads.  Devising
an algorithmic selection of optimal work group size thus becomes an exhaustive
search.  Note that selecting the wrong work group size may slow down execution
by 5--7 times on Adreno GPUs. 

Despite these challenges, we conducted extensive investigations into optimizing
the work group size, focusing primarily on \textsc{conv\_2d} and
\textsc{depthwise\_conv}, as these make up nearly 90\% of the workload for
convolutional networks.  While the algorithmic solution is not perfect, the
alternative brute-force approach is impractical for real time applications
because the work group investigation for a model may take several minutes. In
addition, measurements may be inconsistent due to device temperature, resource
racing, etc., causing the true global optimal work group size to change from one
inference to another. 

Because of these fluctuations, we approximate a reasonable optimum within the
neighborhood region of the global optimum given an inference time function $T(W,
C)$, where $W$ is work group sizes, and $C$ identifies convolution
configuration.  The domain of the function parameters are:
\begin{itemize}
\setlength{\parskip}{0pt}
\setlength{\itemsep}{0pt plus 1pt}
\item Work groups dimensions $W$: $2$, $4$, or $8$
\item Convolution configurations $C$ search space:
\begin{list}{$\circ$}{}
  \setlength{\parskip}{0pt}
  \setlength{\itemsep}{0pt plus 1pt}
  \item \textsc{conv\_2d} weights $1{\times}1$, $2{\times}2$, $3{\times}3$, or 
  \item \textsc{depthwise\_conv} input and output shapes from $(8,8,8)$ to $(128,128,128)$, and
  \item Strides $1{\times}1$, $2{\times}2$, $3{\times}3$
\end{list}
\end{itemize}
Given the search space defined by the convolution configuration, a gradient
descent approach allows us to converge on a stable optimum work groups where
expected performance varies $10\%$ on every inference. From this region of
stable work groups, an approximate optimal work group can be selected for every
device and convolution type combination. 

Work groups from the Table \ref{tab:optimal_wgs} are currently used in TFLite GPU
and their stability is statistically proven.  While they do not necessarily result
in peak optimal time across all parameters,  they are reliable in giving top 10\%
performance regardless of the convolution parameters.

\begin{table}[!h]
\begin{center}
\begin{tabular}{|l|c|c|}
\hline
Adreno GPU Model & \textsc{conv\_2d} & \textsc{depthwise\_conv} \\
\hline\hline
630                 & $(4,8,4)$      & $(4, 4, 8)$ \\
540                 & $(8,2,2)$      & $(8, 8, 2)$ \\
510                 & $(8,4,4)$      & $(8, 4, 4)$ \\
509                 & $(8,4,8)$      & $(8, 4, 2)$ \\
50X/4XX             & $(8,4,8)$      & $(8, 4, 8)$ \\ 
\hline
\end{tabular}
\end{center}
\caption{Optimal work group sizes for Adreno GPUs.}
\label{tab:optimal_wgs}
\end{table}

\section{Memory Manager for Intermediate Tensors}

While we allocate GPU memory for all input/output tensors and tensors holding
the trained weights, we do not allocate memory for all intermediate tensors
between the operators separately, as they do not have to co-exist in memory
simultaneously.  This is an important optimization to reduce the memory
footprint of the GPU run-time.

During initialization, we first topologically sort the network to determine the
execution order of each operator, and the correspondingly required tensors.  For
each intermediate tensor, we can determine the first and the last operator that
uses this tensor either as input or output.  Once the last ``consumer'' of an
intermediate tensor has finished executing, the memory for the said intermediate
tensor can be re-used for other intermediate tensors. To minimize the total
required memory allocation, we have devised a strategy to determine when this
final operator execution has occurred.  This problem is
NP-complete~\cite{sethi1975complete}.

We compared three algorithms for managing the intermediate tensors: (a) a
na\"ive algorithm, (b) a greedy algorithm, and (c) a minimum-cost flow
algorithm.  The first just na\"ively allocates all memory necessary and only
serves as a baseline for comparison.  The latter two implement smart memory
management and use the concept of ``shared objects'' by which we refer to as
allocated memory that is used for more than one tensor during inference, but not
more than exactly one at a time.  The size of the shared object is the maximum
of sizes of tensors that it is used for.  For example, if a shared object $S$ is
used for tensor $a$, re-used for tensor $b$, and later for tensor $c$, the size
of the shared object $S$ needs to be $\mathit{size_S} = \max(\mathit{size_a},
\mathit{size_b}, \mathit{size_c})$.

\begin{algorithm}[!b]
\caption{Greedy Memory Management}
\label{alg:greedy}
\begin{algorithmic}[1]
\STATE $\mathit{available\_objects} \leftarrow \emptyset$
\STATE $\mathit{used\_objects} \leftarrow \emptyset$
\STATE \textbf{for each} $\mathit{op} \in \mathit{operators}$ \textbf{do}

\STATE \hspace{0.3cm}\textbf{for each} $t \in \mathit{op.outputs}$ \textbf{do}
\STATE \hspace{0.6cm}\textbf{if} $t$ is intermediate \textbf{then}
\STATE \hspace{0.9cm}\textbf{if} $\mathit{available\_objects} = \emptyset$ \textbf{then}
\STATE \hspace{1.2cm}$S \leftarrow$ new shared object with size $\mathit{t.size}$
\STATE \hspace{0.9cm}\textbf{else}
\STATE \hspace{1.2cm}$S \leftarrow \mathit{available\_objects}$.find($\mathit{t.size}$)
\STATE \hspace{1.2cm}$\mathit{available\_objects}$.remove($S$)
\STATE \hspace{1.2cm}\textbf{if} $\mathit{t.size} > \mathit{S.size}$ \textbf{then}
\STATE \hspace{1.5cm}$\mathit{S.size} \leftarrow \mathit{t.size}$
\STATE \hspace{0.9cm}$\mathit{t.shared\_object} \leftarrow S$
\STATE \hspace{0.9cm}$\mathit{used\_objects}$.insert($S$)

\STATE \hspace{0.3cm}\textbf{for each} $t \in \mathit{op.inputs}$ \textbf{do}
\STATE \hspace{0.5cm}\textbf{if} $t$ is intermediate \AND $op$ is its last consumer \textbf{then}
\STATE \hspace{0.9cm}$S \leftarrow \mathit{t.shared\_object}$
\STATE \hspace{0.9cm}$\mathit{used\_objects}$.remove($S$)
\STATE \hspace{0.9cm}$\mathit{available\_objects}$.insert($S$)
\end{algorithmic}
\end{algorithm}

\vspace{-\topsep}
\paragraph{The Greedy Algorithm} is summarized in Algorithm~\ref{alg:greedy}.
We iterate through all operators in topological execution order.  If an output
tensor of the current operator is an intermediate tensor, it is assigned to a
newly created shared object if the pool of shared objects is empty (L.7), or to
an existing shared object that has the closest size by absolute difference to
the $\mathit{t.size}$ (L.9) which gets removed from the available pool (L.10).
If $\mathit{t.size} > \mathit{S.size}$, then the shared object's buffer size is
increased (L.11--12).  This shared object $S$ is inserted into the set of
currently used objects (L.14).  After the output tensors, the input tensors are
inspected.  If an input tensor is an intermediate tensor and the current
operator is the last consumer, we remove the shared object that is assigned to
this tensor from the set of currently used objects, and add it back to the pool
of shared objects (L.17--19).

\begin{figure}[b]
\begin{center}
   \includegraphics[height=1.6cm]{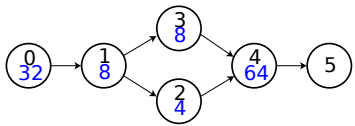}
\end{center}
\caption{
  An example neural net. Each vertex corresponds to an op.  The upper number
  denotes the execution order, and the lower number the size of its output
  intermediate tensor.  The last op does not have the latter as its output
  is not an intermediate tensor.}
\label{fig:graph_example}
\end{figure}

This algorithm has the runtime complexity of $O(n\log{n})$ where $n$ is the
number of intermediate tensors.  We use binary search tree for the pool of
shared objects and binary heap priority queue for the set of currently used
objects.  Straightforward implementation of the same algorithm without these
data structures has a run-time complexity of $O(n^2)$.  For the neural network
from Figure~\ref{fig:graph_example}, this approach re-uses memory of output
tensor of vertex 0 for output tensor of vertex 2, and memory of output tensor of
vertex 1 for output tensor of vertex 4.  The total size of allocated memory is
104.

\vspace{-\topsep}
\paragraph{The Minimum-Cost Flow Algorithm} involves creating an auxiliary flow
network and solving the minimum-cost flow problem (MCFP)~\cite{mcfp}.  First, we
insert two vertices for each intermediate tensor $x$ and denote them $l_x$ and
$r_x$ with two special vertices for the source $s$ and the sink $t$.  Then, we
add directed edges to the flow network:
\begin{enumerate}
\setlength{\parskip}{0pt}
\setlength{\itemsep}{0pt plus 1pt}
\item For each $x$ in $1..N$, add an edge from $s$ to $r_x$ with capacity $1$ and cost $\mathit{size_x}$.
      For tensor $x$, we can allocate new shared object of size $\mathit{size_x}$.
\item If a shared object allocated for tensor $x$ can be re-used for tensor $y$, then add an edge from $l_x$ to $r_y$ with capacity $1$ and cost $\max(0, \mathit{size_y} - \mathit{size_x})$.
      If tensor $y$ is greater in size than tensor $x$, we can re-use corresponding shared object, but we might need to allocate $\mathit{size_y} - \mathit{size_x}$ of additional memory.
      This is not always the case, when the shared object can already have a size greater than $\mathit{size_x}$, but it is a good approximation.
\item For each $x$ in $1..N$, add an edge from $s$ to $l_x$ with capacity $1$ and cost $0$.
\item For each $x$ in $1..N$, add an edge from $r_x$ to $t$ with capacity $1$ and cost $0$.
\end{enumerate}

\begin{figure}[t]
\begin{center}
   \includegraphics[width=0.6\linewidth]{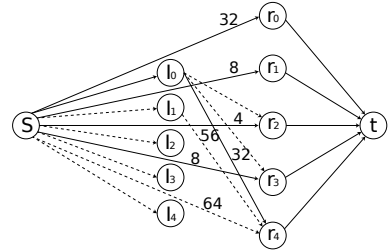}
\end{center}
\caption{
  The flow network for the neural network in Figure~\ref{fig:graph_example}.
  Capacity of each edge is $1$.  Saturated edges,~\ie the final assignment of
  shared objects to tensors, are shown as solid lines.}
\label{fig:mem_algorithm}
\end{figure}

After building the flow network, we solve the MCFP with Shortest Path Faster
Algorithm (SPFA)~\cite{moore1959the} or Johnson's
algorithm~\cite{johnson1977efficient}.  With SPFA, the run-time complexity
$O(N^{4})$, but it can be reduced to $O(N^{3})$ by decreasing the number of
edges of type 2.  Figure~\ref{fig:mem_algorithm} shows a flow network and the
result of this algorithm execution for example graph from
Figure~\ref{fig:graph_example}.  Minimum-cost flow approach re-uses memory of
output tensor of vertex 0 for output tensor of vertex 4.  The total size of
allocated memory is 84.

If an edge of type 1 (from $s$ to $r_x$) is saturated by the flow, \ie its
residual capacity is equal to $0$, we create new shared object for the tensor
$x$. If an edge of type 2 (from $l_x$ to $r_y$) is saturated by the flow, we
assign the same shared object for tensor $y$ that was used by tensor $x$.  After
execution of the algorithm, the amount of the flow will be equal to $N$. It
means that the resulting flow network has information about the assignment of
shared objects for all $N$ intermediate tensors. Size of each shared object is
determined by the maximum size of all tensors assigned to it.

There is no clear winner between these two memory management algorithms
in terms of the minimal memory footprint, and it depends on the network
(Table~\ref{tab:mm}). TFLite GPU is using the greedy algorithm by default
with the developer being able to choose the MCFP algorithm if desired.
\begin{table}[!b]
\begin{center}
\begin{tabular}{|l|c|c|c|}
\hline
Strategy & MobileNet & MobileNetV2 & DeeplabV3\\
\hline\hline
Na\"ive & 9.6 & 13.2 & 24.3 \\
Greedy  & \textbf{2.3} &  4.0 &  \textbf{3.6} \\
MCFP    & 2.7 &  \textbf{3.8} &  4.2 \\
\hline
\end{tabular}
\end{center}
\caption{Total memory allocated (in MB) for all intermediate tensors.
Na\"ive means no memory manager and serves as baseline.
Bold number means the smallest memory footprint for each model.}
\label{tab:mm}
\end{table}
 
\section{Results}

\begin{figure}[t]
\begin{center}
   \includegraphics[width=\columnwidth]{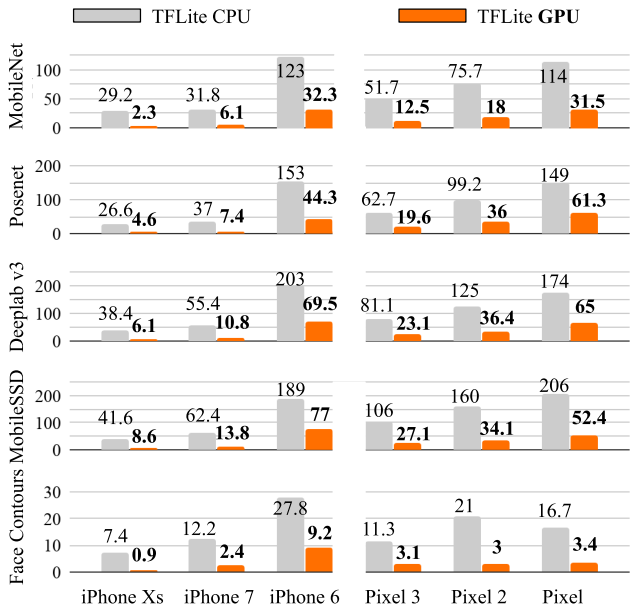}
\end{center}
\caption{
  Average inference latency (in milliseconds) of TFLite GPU (orange) compared to
  CPU (gray) on various neural networks, run on a variety of smartphones (best
  viewed in color).
}
\label{fig:cpu_vs_gpu_timings}
\end{figure}

Figure~\ref{fig:cpu_vs_gpu_timings} illustrates the performance of GPU inference
compared to CPU inference in TFLite for various neural networks which generally
demonstrates a 2--9$\times$ speedup.  The first 10 warm-up runs were skipped for
benchmarking and averages are based on the 100 subsequent inferences.  This
profiling revealed that TFLite GPU is often bound by memory bandwidth and we
typically only see 20--40\% ALU utilization.  On iOS devices, we benefit from
larger cache sizes that result in reduced memory I/O latency, and hence, better
performance than the OpenGL backend.

Table \ref{tab:ios} and Table \ref{tab:android} show the average inference
latency of iOS- and Android-compatible ML frameworks on MobileNet v1,
respectively.  Note that TFLite GPU employs OpenGL for the widest coverage with
reasonable performance.  MACE and SNPE employ OpenCL and may outperform TFLite
GPU on some mobile devices shipped with OpenCL.  As OpenCL is not a part of the
standard Android distribution, apps using those frameworks may not be able to
guarantee their inference performance \eg on Google Pixel devices.  Also note
that SNPE does not run on devices with Arm Mali GPUs.

\begin{figure}[t]
\begin{center}
   \includegraphics[width=\columnwidth]{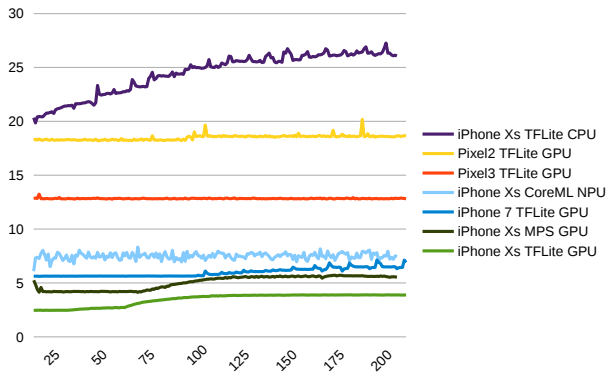}
\end{center}
\caption{
  Inference latency (in milliseconds) for MobileNet v1 over extended period of
  time $[0, 200]sec$ (best viewed in color).
}
\label{fig:latency_over_runtime}
\end{figure}

Figure~\ref{fig:latency_over_runtime} shows how inference performance degrades
over a sustained period of time due thermal throttling of the device. Mobile
inference by applications typically occur in one of two modes: one-time
detection or ongoing run-time data processing. For one-time inference,~\eg
object detection, an application may achieve the peak performance illustrated in
the left half of graph in Figure~\ref{fig:latency_over_runtime} where device
temperature is nominal.  For ongoing run-time inference, \eg video segmentation,
the right half illustrates the potential impact of thermal throttling due to
sustained performance.

In order to avoid data transfer delays, real-time applications usually place
neural network input/output tensors in a GPU texture or buffer. TFLite GPU
allows using CPU-side tensors as input/output as well. Additionally, CPU-to-GPU
data-transfer efficiency can be controlled via time or power efficient
synchronization mechanisms. The most power-efficient one suspends waiting
threads until the GPU completes its task. The fastest option by comparison,
employs an active spin-lock approach, reducing data acquisition delays by
avoiding operating system process re-scheduling.

\begin{table}[!t]
\begin{center}
\begin{tabular}{|l|c|c|c|}
\hline
iOS Device & TFLite GPU & MPSCNN & CoreML \\
\hline\hline
iPhone Xs & 2.3 & 4.1 & 7.1 \\
iPhone 7  & 5.5 & 7.9 & 42  \\
iPhone 6  & 31  & 92  & 116 \\
\hline
\end{tabular}
\end{center}
\caption{
  Average inference latency (in milliseconds) of iOS-compatible ML frameworks
  on MobileNet v1.}
\label{tab:ios}
\end{table}

\begin{table}[!t]
\begin{center}
\begin{tabular}{|l|c|c|c|}
\hline
Android Device & TFLite GPU & MACE & SNPE \\
\hline\hline
Samsung S9                      & \multirow{2}{*}{13}   & \multirow{2}{*}{12}      & \multirow{2}{*}{6.9} \\
\ \ \ \ {\footnotesize (Adreno 630)}    & & & \\
Xiaomi Mi8 SE                   & \multirow{2}{*}{35.9} & \multirow{2}{*}{29.6}    & \multirow{2}{*}{20} \\
\ \ \ \ {\footnotesize (Adreno 616)}    & & & \\
Huawei P20 Pro                  & \multirow{2}{*}{13.5} & \multirow{2}{*}{45}      & \multirow{2}{*}{N/A$^1$} \\
\ \ \ \ {\footnotesize (Mali G72-MP12)} & & & \\
Google Pixel 2                  & \multirow{2}{*}{18}   & \multirow{2}{*}{N/A$^2$} & \multirow{2}{*}{N/A$^2$} \\
\ \ \ \ {\footnotesize (Adreno 540)}    & & & \\
Google Pixel 3                  & \multirow{2}{*}{12.5} & \multirow{2}{*}{N/A$^2$} & \multirow{2}{*}{N/A$^2$} \\
\ \ \ \ {\footnotesize (Adreno 630)}    & & & \\
\hline
\end{tabular}
\end{center}
\caption{
  Average inference latency (in milliseconds) of Android-compatible ML
  frameworks on MobileNet v1.  Note that TFLite GPU employs OpenGL and thus has
  the widest coverage with reasonable performance.  MACE and SNPE employ OpenCL
  and may run faster on devices shipped with OpenCL, but may not run on all
  devices.  $^1$ Arm Mali GPUs are not compatible with SNPE.  $^2$ Google Pixel
  devices do not support OpenCL.}
\label{tab:android}
\end{table}

\section{Conclusion}

In this paper, we presented the architectural design of TFLite GPU.  We
described the properties of mobile GPUs and explained optimization techniques we
employed for fast memory I/O, small run-time memory footprint, and fast compute
shader execution.  With these, we aim to make the network architects be mobile
GPU-aware when they design their networks.

From our discussion of mobile GPU-friendly data layout PHWC4, neural network
designers should know that any kind of \textsc{reshape}s are significantly more
expensive on the GPU than on the CPU.  The network itself will learn the weights
regardless of the \textsc{reshape} op, thus it is best to skip the operator
entirely if a \textsc{reshape} operation was inserted just for convenience of
the architect.

For the same reason, if the mobile device can produce RGBA rather than RGB, it
is now apparent that using the former can avoid a conversion, \ie memory copy,
from RGBA to RGB.  Similarly, if the mobile device can render a 4-channel
tensor, \ie RGBA, directly, that can be a better choice than the RGB
counterpart.  This choices benefits not just the graph input/output, but also
its intermediate tensors.  Similarly, since we know that a tensor of shape
$[B,H,W,5]$, for instance, is twice as expensive as $[B,H,W,4]$, but about the
same as $[B,H,W,8]$, then the architect can tune around those 4-channel
boundaries rather than trying to optimize on other boundaries.

TFLite GPU is still in its early development stages.  We plan to investigate
several areas including employing additional GPU-specific optimizations to
improve inference speed further, and expanding support for more operations,
\eg understand more about recurring networks or LSTMs, and how we can optimize
those for GPUs.  Finally, we are extensively exploring other GPU backends such
as OpenCL and Vulkan to achieve better ALU utilization.

\section*{Acknowledgements}

We would like to acknowledge our colleagues at TensorFlow Lite; Lawrence Chan,
Tim Davis, Jared Duke, Yu-Cheng Ling, Andrew Selle, Sarah Sirajuddin, and Pete
Warden.  We are also grateful to Aleksandr Ignashev for the figures in this
paper and Karthik Raveendran for his valuable feedback.

{\small
\bibliographystyle{ieee_fullname}
\bibliography{egbib}
}

\end{document}